\documentclass{article}
\usepackage{ijcai21}

\usepackage{times}

\usepackage{soul}
\usepackage{url}
\usepackage[hidelinks]{hyperref}
\usepackage[utf8]{inputenc}
\usepackage[small]{caption}
\usepackage{amsmath}
\usepackage{booktabs}
\usepackage{bm}
\usepackage{enumerate}
\usepackage{amsfonts,amssymb}
\usepackage{color}
\usepackage{graphicx}
\usepackage{subfigure}
\usepackage{diagbox}
\usepackage{algorithm}
\usepackage{algorithmic}
\usepackage{setspace}

\usepackage{multirow}

\usepackage{hyperref}

\usepackage{amsthm}

\title{A Coalition Formation Game Approach for Personalized Federated Learning} 

\author{
Leijie Wu$^1$\footnote{Contact Author}\and
Song Guo$^1$\and
Yaohong Ding$^1$\and
Yufeng Zhan$^2$\And
Jie Zhang$^1$\\
\affiliations
$^1$The Hong Kong Polytechnic University\\
$^2$Beijing Institute of Technology\\
\emails
lei-jie.wu@connect.polyu.hk, song.guo@polyu.edu.hk, yaohong.ding@connect.polyu.hk, yu-feng.zhan@bit.edu.cn, 18104473r@connect.polyu.hk
}

\begin{document}

\maketitle

\begin{abstract} 
\label{Sec:abs}
Facing the challenge of statistical diversity in client local data distribution, personalized federated learning (PFL) has become a growing research hotspot.
Although the state-of-the-art methods with model similarity based pairwise collaboration have achieved promising performance, they neglect the fact that model aggregation is essentially a collaboration process within the coalition, where the complex multiwise influences take place among clients.
%
In this paper, we first apply Shapley value (SV) from coalition game theory into the PFL scenario.
To measure the multiwise collaboration among a group of clients on the personalized learning performance, SV takes their marginal contribution to the final result as a metric.
We propose a novel personalized algorithm: pFedSV, 
which can 1.~identify each client's optimal collaborator coalition and 2.~perform personalized model aggregation based on SV.
Extensive experiments on various datasets (MNIST, Fashion-MNIST, and CIFAR-10) are conducted with different Non-IID data settings (Pathological and Dirichlet).
The results show that pFedSV 
can achieve superior personalized accuracy for each client, compared to the state-of-the-art benchmarks.
\end{abstract}

\section{Introduction}
\label{Sec:intro}

Federated learning (FL) is a recent  promising distributed machine learning technique, which can collaboratively train a shared model among multiple clients with data privacy protection \cite{mcmahan2017communication}.
%
%
%
The effectiveness of this shared-model scheme highly depends on similar local data distribution among clients, which is called independent and identically distribution (IID).
However, in the vast majority of real-world scenarios, the data distribution of clients is Non-IID with significant heterogeneity, also called \textit{statistical diversity}.
This phenomenon is particularly evident in the cross-silo FL \cite{kairouz2019advances}, where even the label distribution of each client is distinctly different.
%

To deal with the above challenges, some researchers have noted that statistical diversity may be an advantage when viewing the issue from the perspective of personalized federated learning (PFL).
Initially, some methods try to add an extra fine-tuning step to the trained shared-model by only using local data \cite{cortes2014domain,mansour2020three,wang2019federated}.
In a further step, to facilitate cooperation between clients for mutual progress, they encourage collaboration between clients based on their pairwise model similarity (or loss differences), which determine the respective aggregation weights \cite{huang2021personalized,zhang2020personalized,collins2021exploiting}. 
In this way, clients with similar model features will make more contribution in the collaboration.

\begin{figure}[t]
\centering
\includegraphics[width=0.46\textwidth]{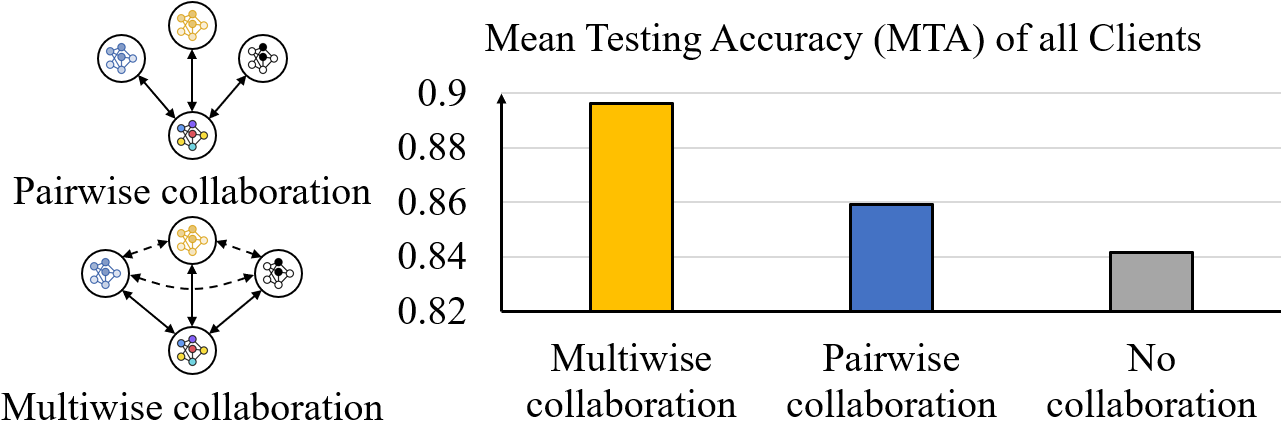}
\caption{The schematic of Multiwise vs. Pairwise collaboration and the pre-experimental results on CIFAR-10 dataset with the pathological Non-IID setting.}
\label{Fig:PreEx Multiwise vs. Pairwise}
\vspace{-7pt}
\end{figure}


However, latest research reveals the fact that model aggregation is essentially a coalition game \cite{donahue2021model}.
Thus, the collaboration within the coalition needs to consider the multiwise influences among clients.
%
In Fig.~\ref{Fig:PreEx Multiwise vs. Pairwise}, we illustrate the schematic of Multiwise vs. Pairwise collaboration and the pre-experimental results, which empirically prove the superiority of multiwise collaboration.
More specifically, we argue that pairwise collaboration for PFL cannot address two critical challenges, which will be elaborated later in Sec. \ref{Sec:problem_root} with detailed experiments analysis.
\textbf{First is data relevance}, each client actually wants to collaborate with others whose local data distribution is relevant to themselves, but the simple one-to-one model parameter similarity testing cannot guarantee clients with similar models will also have higher relevance on their local data, since they ignore the fact that client relevance needs to be analyzed in a multiwise collaboration.
%
\textbf{Second is multiwise aggregation weights}, when performing personalized model aggregation, the collaboration within the client coalition must consider their multiwise influences on the final result, while the aggregation weights in pairwise collaboration only rely on myopia one-to-one model similarity.
%

Carrying the above insights, in this paper, 
we are the first to dissect PFL problems via the lens of coalition game theory \cite{myerson2013game}. 
We introduce \textbf{Shapley Value (SV)} from coalition game theory to evaluate each client's marginal contribution to the personalized performance, which is naturally based on multiwise influences analysis within the coalition. 
%
Specifically, the desirable properties of SV ensure us to simultaneously address above critical issues with a multiwise collaboration solution.
Our key technical contributions are:
\begin{itemize}
\item As far as we know, we first provide an in-depth analysis to PFL scenario via the lens of coalition game theory. We introduce SV to evaluate each client's marginal contribution during the personalization process, which is based on multiwise influences analysis within the collaborator coalition.

\item We propose a novel pFedSV algorithm that exploits the unique properties of SVs to simultaneously address the key issues in PFL, where the \textit{Null player} and \textit{Symmetry} properties are applied to identify data-relevant client coalition, and the \textit{Linearity} and \textit{Group rationality} properties are employed for each client's personalized model aggregation with multiwise collaboration.

\item Eventually, we conduct extensive experiments to evaluate the performance of pFedSV on several realistic datasets with different Non-IID data settings. The results show that pFedSV can outperform the state-of-the-art methods in the personalized accuracy of each client.

\end{itemize}

\section{Related Work}
\label{related_work}



Recently, to address the statistical diversity challenge of clients with Non-IID data, Personalized Federated Learning (PFL) has emerged as a solution scenario that is attracting more attention.
Initially, additional fine-tuning step on the client's local dataset is a natural strategy for personalization \cite{mansour2020three,wang2019federated}, and some prior studies have attempted to enhance the robustness of global model under severe non-IID level by regularization \cite{t2020personalized} or add a proximal term \cite{li2020federated}.
%
However, they are all adjusted on single global model scheme which cannot satisfy the personalized demand of individual clients at the local data level, as the target distribution of clients in severe Non-IID setting can be fairly different from the global average aggregation \cite{jiang2019improving}.

With the above challenges, some recent methodes consider to train a personalized model for each client that perfectly adapts to their local targets. pFedHN employs a hypernetwork to directly generate personalized parameters for each client's model \cite{shamsian2021personalized}.
To promote cooperation between clients with relevant local target distributions to achieve mutual progress, FedFomo \cite{zhang2020personalized} and FedAMP \cite{huang2021personalized} encourage pairwise collaboration for clients with similar model features, where the former uses loss similarity and the latter adopts parameter similarity.

Although pairwise collaboration methods have achieved good results, they ignore the fact that model aggregation is a coalition game, which requires considering the multiwise influences among collaborators. 
Different from existing work, we introduce SV from coalition game theory to analyze the multiwise influences among clients, which is based on their marginal contribution to the personalization of others.


\section{The Essence of PFL Problem}
\label{Sec:essence of PFL}
In this section, we first introduce the objectives of PFL and the corresponding problem formulation \cite{kairouz2019advances,zhao2018federated}. Then, we will delve into the root causes of the problems through various pre-experiments analysis and present our multiwise collaboration solution: Shapley value, to address these problems.

\subsection{Problem Formulation}
\label{Sec:problem_formulation}
PFL aims to customize personalized models for each client to accommodate their private data distribution through collaboration between a set of clients. 
Considering $n$ clients $C_1, C_2, \cdots, C_n$ with the same structure of model $\mathcal{M}$ but parameterized by different weights $\theta_1, \theta_2, \cdots, \theta_n$, their respective personalized models can be denoted by $\mathcal{M}(\theta_i)$. 
Unlike traditional federated learning, the private dataset $\mathcal{D}_i$ of each client $i$ is uniformly sampled from their own distinct data distribution $\mathcal{P}_i$. 
Let $\ell_i$ denote the corresponding loss function for client $i$, and $\mathcal{L}_i$ the average loss over the private dataset $\mathcal{D}_i$ is denoted by $\mathcal{L}_i(\theta_i) = \frac{1}{d_i} \sum_{j\in \mathcal{D}_i} \ell_i(j,\theta_i)$, where $d_i$ is the data size of $\mathcal{D}_i$ and $j$ is one of the data samples in $\mathcal{D}_i$.
The optimization objective of PFL is
\begin{equation}
    \Theta^* = \arg\min_\Theta \frac{1}{n} \sum_{i=1}^n \mathcal{L}_i(\theta_i),
\end{equation}
where $\Theta$ is the set of personalized model parameter $\{ \theta_i \}_{i=1}^n$.

\begin{figure}[t]
\centering
\includegraphics[width=0.46\textwidth]{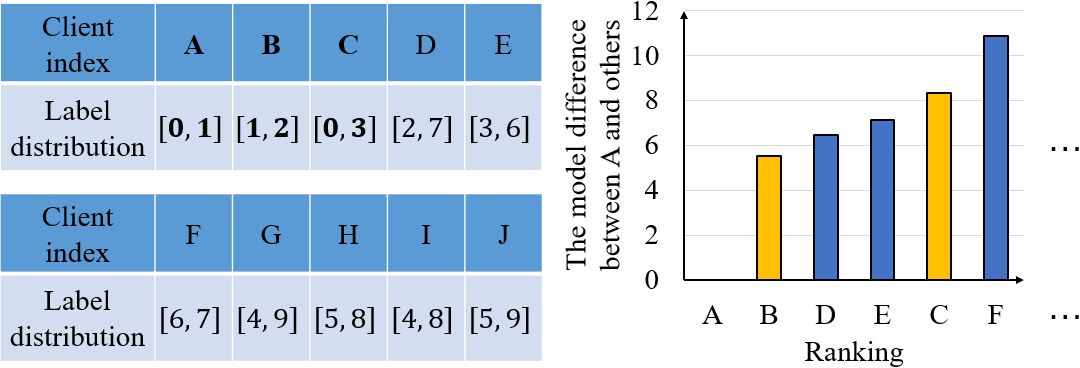}
\caption{Pre-experiment on CIFAR-10 with the pathological Non-IID setting, where the table shows the details of client data distribution and the bar chart shows the model different $||\theta_A-\theta_i||^2$ between client $A$ and other $i\in \{N\}$.}
\label{Fig:PreEx_similarity_new}
\vspace{-7pt}
\end{figure}

\subsection{Root Causes of PFL Problems}
\label{Sec:problem_root}

\textbf{Data Relevance.} According to extensive previous work for data Non-IID in FL \cite{zhao2018federated,li2020federated,li2019convergence,karimireddy2020scaffold}, the model performance degradation in the Non-IID case is due to the significant local level data distribution differences among clients.
By the same token, for better model personalization in PFL scenario, each client should seek to collaborate with others whose local data distribution is truly relevant to their own. 
In the table of Fig.~\ref{Fig:PreEx_similarity_new}, we explain what is data relevance using the MNIST dataset (ten labeled digits from $0$ to $9$) with pathological Non-IID, i.e., each client randomly has two types of labels with equal data size. 
%
Take client $A$ with labels $[0,1]$ as an example, client $B$ with labels $[1,2]$ and client $C$ with labels $[0,3]$ are its data-relevant clients, which are the clients that $A$ really wants to collaborate with in its own model personalization process.
However, the one-to-one model similarity testing is myopia and ineffective as it completely ignores the multiwise influences among clients.
%
Still using client $A$ as an example, we adopt $||\theta_A-\theta_i||^2, i\in \{N\}$ to measure the model parameter difference between $A$ and other models. 
If the model similarity theory is true, the model differences of \textit{B} and \textit{C} should be the smallest among all clients, which is contrary to our experiment results in Fig.~\ref{Fig:PreEx_similarity_new}.


\textbf{ Multiwise Collaboration Weights.}
Another key-point in PFL is the personalized model aggregation targeting each client's local data level features.
%
%
The previous methods adopted pairwise collaboration by comparing model similarities one-to-one and assigning proportional aggregation weights based on their magnitudes, which is demonstrated in Fig.~\ref{Fig:PreEx Multiwise vs. Pairwise}.
%
%
However, imagine a scenario where the client's current model is a carriage, and every other client's model is a force that moves the carriage in a certain direction, and the destination of the carriage is the client's optimal personalized model.
Obviously, the movement of carriage is the result of multiple forces combination, which indicates that the multiwise influences among collaborators must be considered when generating the personalized model aggregation weights.
%
Under the same conditions that their respective data-related customers are informed in advance, we conduct extensive pre-experiments where the only variable is the collaboration methods among clients when generating aggregated weights. The results in Fig.~\ref{Fig:PreEx Multiwise vs. Pairwise} indicating that multiwise collaboration outperforms pairwise collaboration.


\subsection{ The Coalition Game Solution: Shapley Value }
\label{Sec:problem_solution_sv}

The SV from coalition game theory help us to evaluate each client's marginal contribution to the personalization of others, its calculation process involves the performance analysis of clients in different combinations, which naturally incorporates the multiwise influences analysis we need within the coalition. 
Therefore, we adopt SV as our multiwise collaboration solution to address two above critical challenges.


\textbf{Preliminaries of SV.} 
Consider each client as a player in the coalition game, where $N=\{1,2,\cdots,n\}$ denotes the set of players. A \textit{utility function} $v(S): 2^n \to \mathbb{R}$ assigns to every coalition $S \subseteq N$ a real number representing the gain obtained by the coalition as a whole. By convention, we assume that $v(\emptyset)=0$. 
Formally, let $\pi\in \Pi(N)$ denote a permutation of clients in $N$, and $C_{\pi}(i) = \{ j\in\pi: \pi(j)<\pi(i) \}$ is a coalition containing all predecessors of client $i$ in $\pi$.
The SV of client $i$ is defined as the average marginal contribution to all possible coalitions $C_\pi(i)$ formed by other clients:
\begin{equation}
\label{Eq:sv_calculation}
    \varphi_i(v) = \frac{1}{|N|!} \sum_{\pi\in\Pi}[v(C_\pi(i)\cup\{i\})-v(C_\pi(i))].
\end{equation}
The formula in (\ref{Eq:sv_calculation}) can also be rewritten as:
\begin{equation}
\label{Eq:sv_calculation_rewritten}
    \varphi_i(v) = \sum_{S\in N\setminus \{i\}} \frac{|S|!(|N|-|S|-1)!}{|N|!} [v(S\cup\{i\})-v(S)].
\end{equation}

The importance of SV in addressing the two root problems stems from its desirable and unique properties as follows.




\textbf{SV for Data Relevance.}

\begin{itemize}
    \item \textit{Symmetry}: Two clients who have the same contribution to the coalition should have the same value. That is, if client $i$ and $j$ are equivalent in the sense of $v(S\cup\{i\})=v(S\cup\{j\}), \forall S \subseteq N \setminus \{i,j\}$, then $\varphi_i = \varphi_j$.
    \item \textit{Null Player}: Client with zero marginal contributions to all possible coalitions is null player and receive zero payoff, i.e., $\varphi_i=0$ if $v(S\cup\{i\}) = 0$ for all $S \subseteq N \setminus \{i\}$.
\end{itemize}

The \textit{Symmetry} and \textit{Null Player} properties in SV can assist us in precisely identifying their data-relevant clients, where those irrelevant clients will be identified as \textit{null} or even \textit{negative}. 
%
For PFL scenario, in each communication round $t$, each client $i$ will first upload their local updated model parameter $\theta^t_i$ to the server, forming a model pool $\{\theta_i^t\}_{i=1}^n$ on the server-side. 
Then, they will download the relevant clients' model parameters from the model pool for their personalization. 
Of course, at the beginning, they must have no idea who are their data-relevant clients in the model pool. 
Therefore, we construct a model download vector for each client based on the relevance score, which can guarantee their data-relevant clients will be identified within a certain rounds .

At the beginning, for each client $i, i \in N$, it generates an $n$-dimensional all-zero relevance vector $\phi^{i,t}$, where $\phi_{j}^{i,t}$ denotes the relevance score of client $j$ to $i$ in $t$-th round and we have $\phi^{i,t=1} = \Vec{0}$. 
Thus initially it will randomly download $k$ other model parameters, and later it will choose to download the top-$k$ with non-negative relevance. Then, we form a set $S_{i,k}^t$ with its own and downloaded $k$ model parameters.
Now, client $i$ can compute the relevance score of the downloaded models by using the following coalition game and the local validation dataset $\mathcal{D}_{V_i}$. 
We define a coalition game $(\{\theta_j^t\}_{j\in S_{i,k}^t}, v)$, where $v$ is a utility function that assigns a value to each $X\subseteq S_{i,k}^t$. 
Here, we define the value using the performance $\mathcal{A}$ of the model with parameters $\theta^{t}_X$ generated from $X$ on the validation dataset $\mathcal{D}_{V_i}$ as follows.

\begin{equation}
\label{Eq: define_sv}
    \theta^{t}_X = \frac{1}{|X|}\sum_{j\in X} \theta_j^t, \text{and}\,\,\,  v(X,\mathcal{D}_{V_i}) = \mathcal{A}(\theta^{t}_X, \mathcal{D}_{V_i}).
\end{equation}
Then, we can obtain the SV $\varphi_j^t, j\in S_{i,k}^t$ of all downloaded model parameters from the coalition game $(\{\theta_j^t\}_{j\in S_{i,k}^t}, v)$ in $t$-th round according to Eq.~(\ref{Eq:sv_calculation}).
Next, client $i$ will updated its relevance vector to $\phi^{i,t+1}$ as below:
\begin{equation}
\label{Eq:relevance_update}
    \phi^{i,t+1}_j = \alpha\phi^{i,t}_j +(1-\alpha)\varphi_j^t, \forall j\in S_{i,k}^t.
\end{equation}

Intuitively, a larger relevance score for client $j$ means that it contributes more to the personalized performance of client $i$, and therefore has a higher likelihood of being the data-relevant client.
Besides, we notice that the relevance vector is unstable in the initial stage and requires several rounds of iterative updates.
However, by definition, when other clients' model parameters negatively affect the personalized performance in the coalition game, its SV can be negative, so the irrelevant clients' scores will rapidly decrease in the iterations and thus excluded. 
For example, if client $i$ has $2$ data-relevant clients in total $20$ clients and it downloads $5$ other clients' model parameters per round, then it takes at most $5$ rounds to identify all data-relevant clients (The Proof is in Appendix.1).

\textit{Dynamic top-$k$ download mechanism:} The above relevance vector has another crucial role in dynamically adjusting the number of model downloads $k$.
With constant updates, only the score of data-relevant clients can remain non-negative, so when the remaining number does not match the current value of $k$, we dynamically adjust it to ensure that all downloads are for the necessary data-relevant clients.

\begin{algorithm}[!t]
\setstretch{1.0} %
\caption{Shapley value based Personalized federated learning (pFedSV)}
\label{algorthm_pFedSV}
\begin{algorithmic}[1]
    \REQUIRE $n$, $N$, $\{\theta_i\}_{i=1}^n$, $k$, $E$, $T$, $R$ and $\mathcal{D}_{V_i}$.
    \ENSURE $\{\theta_i^*\}_{i=1}^n$: clients' personalized model parameters.
    \STATE Initialize the clients' model parameters $\{\theta_i\}_{i=1}^n$. \label{code1_1}2
    \STATE Initialize clients' relevence vector: $\phi^{i,t=1} = \Vec{0}$, $\forall i \in N$. \label{code1_2}
    \FOR{round $t=1,2,\cdots,T$} \label{code1_3}
        \FOR{client $i=1,2,\cdots,n$} \label{code1_4}
            \STATE update its model parameter to $\theta_i^t$ via $E$ local epochs and upload to the server. \label{code1_5}
            \STATE download $k$ copies of other clients' model parameters from server according to the dynamic top-$k$ download mechanism. \label{code1_6}
            \STATE $S_{i,k}^t \leftarrow \theta_i^t$ $\cup$ \{$k$ downloaded model parameters\}. \label{code1_7}
            \STATE $\varphi_j^t \Leftarrow$ SV\_evaluation($S_{i,k}^t, \mathcal{D}_{V_i}, R$), $\forall j \in S_{i,k}^t$. \label{code1_8}
            \STATE $\phi^{i,t+1}_j = \alpha\phi^{i,t}_j +(1-\alpha)\varphi_j^t, \forall j\in S_{i,k}^t$ \label{code1_9}
            \STATE $w_j^{t*} = \frac{w_j^t}{\sum_j w_j^t}$ $\Leftarrow$  $w_j^t = \frac{\max(\varphi_j^t, 0)}{\lVert \theta_i^t-\theta_j^t \rVert}$, $\forall j \in S_{i,k}^t$. \label{code1_10}
            \STATE $\theta_i^{t*} = \sum_j w_j^{t*} \theta_j^t, \forall j\in S_{i,k}^t$. \label{code1_11}
        \ENDFOR \label{code1_12}
    \ENDFOR \label{code1_13}
\end{algorithmic}
\end{algorithm}

\textbf{SV for Multiwise Collaboration Weights.}

\begin{itemize}
    \item \textit{Group Rationality}: The gain of the entire coalition $S$ is completely distributed among all clients in $S$, i.e., $v(S)=\sum_{i\in S}\varphi_i$.
    \item \textit{Linearity}: The values under multiple utilities sum up to the value under a utility that is the sum of all these utilities: $\varphi_i(v)+\varphi_i(u)=\varphi_i(v+u)$. Also, for every $i \in N$ and any real number $a$, it has $\varphi_i(av) = a\varphi_i(v)$.
\end{itemize}

The \textit{Group  Rationality} and \textit{Linearity} properties perfectly fit the demand of personalized model aggregation with  multiwise collaboration in PFL. 
According to Eq.~(\ref{Eq:sv_calculation}), the computation of SV requires exploring multiple permutations among clients within the coalition game, hence the process naturally takes into account the complex effects of multiwise collaborations among clients. 
Furthermore, the \textit{Group  Rationality} property guarantees that the target of all clients within the coalition is the same, i.e., to achieve the best performance for the current client $i$, which also means the optimal personalized model parameters.
And the \textit{Linearity} property naturally fits into the model aggregation process, i.e., the improvement of personalized accuracy by aggregating other client models into their own is fully reflected in the SV of the model, where a larger positive SV indicates a larger positive contribution to performance improvement and vise verse.

Based on the SV $\varphi_j^t$ for all $j\in S_{i,k}^t$ in Eq.~(\ref{Eq:relevance_update}), the downloaded model parameters are assigned a real number that represents their marginal contribution to the personalization of the current client $i$, where a positive number indicates a positive effect and vice versa. 
Therefore, we first need to exclude those model parameters of irrelevant clients with negative SV out of the multiwise collaboration in current round, and only compute the initial weights for data-relevant clients within the coalition as follows:
\begin{equation}
\label{Eq:initial_weights}
    w_j^t = \frac{\max(\varphi_j^t, 0)}{\lVert \theta_i^t-\theta_j^t \rVert},
\end{equation}
where we adopt the model differences $\lVert \theta_i^t-\theta_j^t \rVert$ to further fine-tune the resulting weights.
Then, we perform $0$-$1$ normalization on the initial weights to obtain their respective aggregation weights $w_j^{t*} = \frac{w_j^t}{\sum_j w_j^t}$, which maintains $w_j^{t*}\in[0,1]$ and $\sum_j w_j^{t*} =1$.
Finally, we generate the personalized model parameters of client $i$ in $t$-th round based on the following multiwise collaboration:
\begin{equation}
\label{Eq:generate_pModel}
    \theta_i^{t*} = \sum_j w_j^{t*} \theta_j^t, \quad  \forall j\in S_{i,k}^t.
\end{equation}
Note that we perform SV evaluations in each round to accommodate small changes in multiwise influences due to parameter changes after client local model training.


\begin{algorithm}[!t]
\setstretch{1.0} %
\caption{Shapley value evaluation}
\label{algorthm_SV_evaluation}
\begin{algorithmic}[1]
    \REQUIRE $S_{i,k}^t$, $\mathcal{D}_{V_i}$, $R$.
    \ENSURE $\varphi_j^t, \forall j\in S_{i,k}^t$.
    \STATE $P \leftarrow$ set of $R$ permutations of $S_{i,k}^t$. \label{code2_1}
        \FOR{client $j\in S_{i,k}^t$} \label{code2_2}
            \FOR{permutation $p \in P$} \label{code2_3}
            \STATE $X_{p,j}^t = \{ l|l\in S_{i,k}^t \land p(l)\leq j \}$  \label{code2_4}
            \STATE $a_j^p \leftarrow v(\{X_{p,j}^t \cup j\}, \mathcal{D}_{V_i}) -  v(X_{p,j}^t, \mathcal{D}_{V_i})$  \label{code2_5}
            \STATE $\varphi_j^t \leftarrow \varphi_j^t + \frac{1}{|P|}a_j^p$. \label{code2_6}
            \ENDFOR
    \ENDFOR
\end{algorithmic}
\end{algorithm}

\begin{figure*}[t]
\centering
\includegraphics[width=\textwidth]{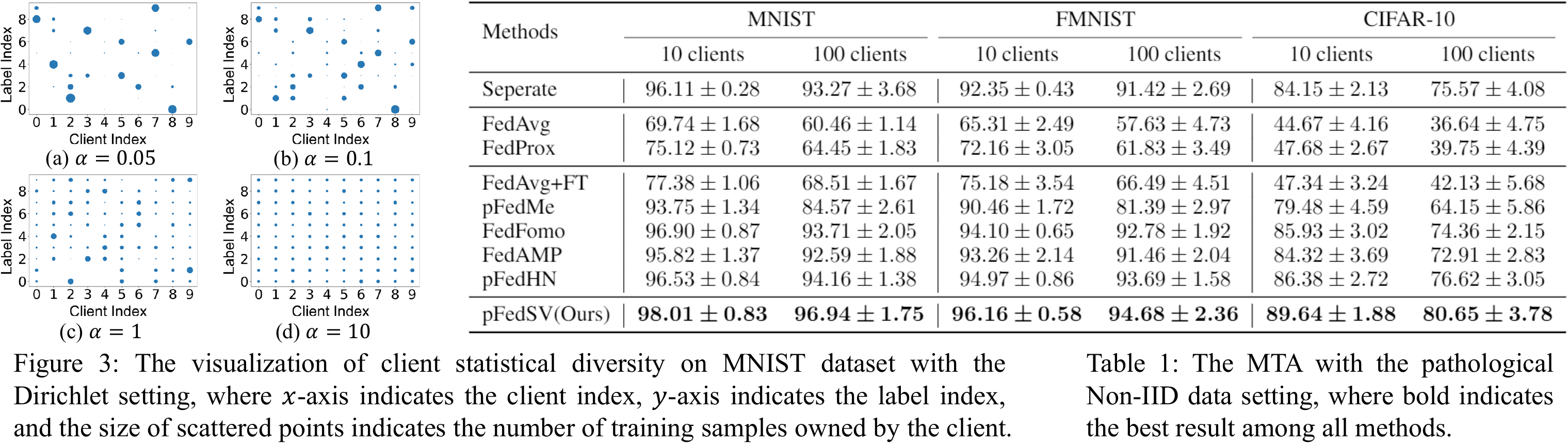}
\end{figure*}

\section{pFedSV Algorithm}
\label{Sec:pFedSV}
Based on the above solution framework, we propose the pFedSV Algorithm shown below. The algorithm \ref{algorthm_pFedSV} demonstrates the personalization process of pfedSV for each client.

In the beginning, each client initialize their model parameters $\theta_i$ and the relevance vector $\phi^{i}$ (Line \ref{code1_1}-\ref{code1_2}). 
Then in each round $t$, they update the model parameters to $\theta_i^t$ by $E$ local epochs training and upload them to the server (Line \ref{code1_5}).
Next, they download $k$ copies of other clients' model parameters according to the dynamic top-$k$ download mechanism (Line \ref{code1_6}). 
At this point, the basic conditions of each client's coalition game for their own model personalization are available.
First, they form a coalition game $(\{\theta_j^t\}_{j\in S_{i,k}^t}, v)$, where $S_{i,k}^t$ is the model parameter set consisting of $k$ downloaded model parameters and their own (Line \ref{code1_7}).
Then, the SV evaluation process is performed to obtain the SV of each model parameters in $S_{i,k}^t$ (Line \ref{code1_8}), which will be elaborated in Algorithm \ref{algorthm_SV_evaluation} later.
Next, the obtained SV are used to address two challenges: updating the relevance vector of each client for identifying their data-relevant clients (Line \ref{code1_9}), and calculating the multiwise aggregation weights for model personalization (Line \ref{code1_10}).
Finally, each client performs the respective weighted aggregation to obtain new model parameters as the starting point for the next round $t+1$.

Since the time complexity required to accurately evaluate SV is exponential to the number of players, we need an approximation algorithm to make the trade-off.
According to Eq.~(\ref{Eq:sv_calculation}), the calculation of SV can be viewed as an expectation calculation problem, so we adopt a widely accepted Monte Carlo sampling technique to approximate the SV \cite{mann1962values,castro2009polynomial,maleki2013bounding}.
The related details are elaborated in Algorithm \ref{algorthm_SV_evaluation}.
First, we randomly sample $R$ permutations of $S_{i,k}^t$ out of total $|S_{i,k}^t|!$ permutations to form a set $P$ (Line \ref{code2_1}).
%
Then, for each permutation, we scan it from the first element to the last and calculate the marginal contribution for every newly added model parameters (Line \ref{code2_3}-\ref{code2_5}).
Perform the same procedure for all $R$ permutations and the approximation of SV is the average of $R$ calculated marginal contributions (Line \ref{code2_6}).
As the number of samples $R$ gradually increases, Monte Carlo sampling will eventually be an unbiased estimate of the SV. Previous work has analyzed error bounds for this approximation and concluded that in practice, $R=3|S_{i,k}^t| \ll |S_{i,k}^t|!$ Monte Carlo samples is sufficient for convergence \cite{maleki2013bounding}.

\begin{figure*}[t]
\centering
\includegraphics[width=\textwidth]{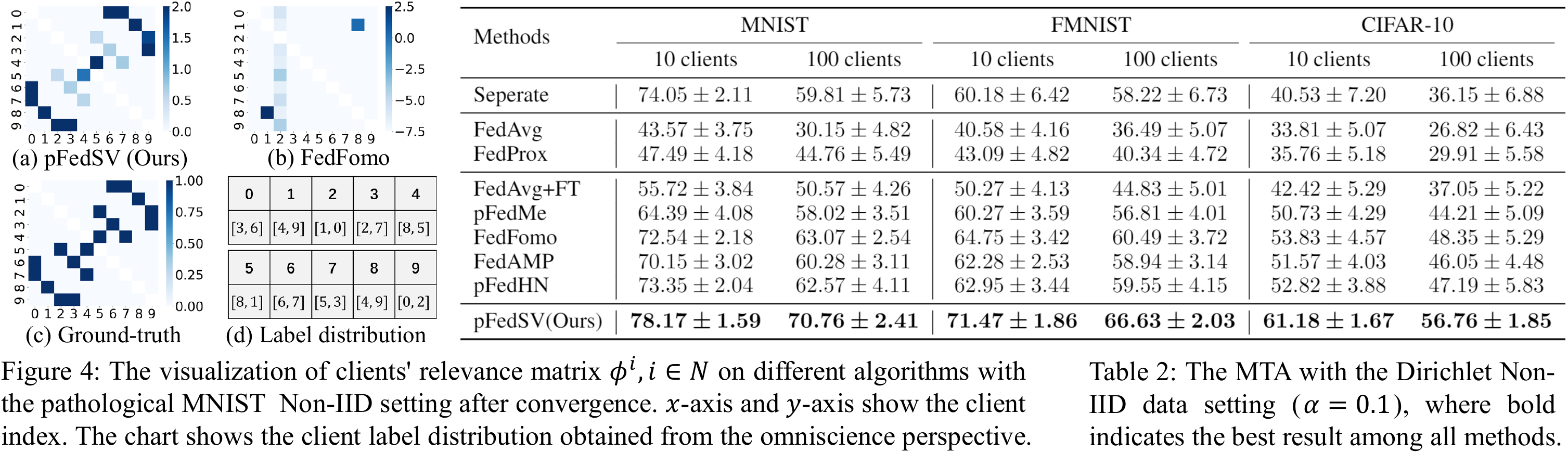}
\end{figure*}

\section{Experiments}
\label{Experiment}

\subsection{Experimental Setup}
\label{Experimental Setup}

In this section we will show all the experiment setup, including hyperparameter settings, datasets, baselines, etc.

\textbf{Baselines \& Evaluation Metric.} 
We evaluate the performance of pFedSV by comparing it with the state-of-the-art PFL algorithms, including pFedMe \cite{t2020personalized}, pFedHN\cite{shamsian2021personalized}, FedFomo \cite{zhang2020personalized},and FedAMP\cite{huang2021personalized}.
For a more comprehensive understanding, we also add the classical single global model methods FedAvg \cite{mcmahan2017communication}, FedAvg with fine-tuning (FedAvg+FT) and FedProx \cite{li2020federated}, as well as the simplest separate local training named \textit{separate}, where each client individually train their own model without collaboration.
The performance of all algorithms is evaluated by the mean testing accuracy (MTA), which is the average of the testing accuracy on all clients and the $\pm$ indicates the error range of the MTA after $5$ repeated experiments.

\textbf{Non-IID data setting.}
We conduct experiments on three public benchmark datasets, MNIST \cite{lecun1998mnist}, FMNIST (Fashion-MNIST) \cite{xiao2017fashion} and CIFAR-10 \cite{krizhevsky2009learning}. 
For each dataset, we adopt two different Non-IID settings: 
I. The pathological Non-IID data setting that each client is randomly assigned two types of labels and the privacy data is not similar between any two clients, which is shown as Fig.~\ref{Fig:PreEx Multiwise vs. Pairwise}.
II. The Dirichlet Non-IID data setting \textbf{Dir}($\alpha$), where a smaller $\alpha$ means higher data heterogeneity, as it makes the client label distribution more biased. We visualized the effect of different $\alpha$ on clients' statistical diversity for MNIST dataset in Fig.~3.

\textbf{Implementation details.}
We consider two FL scenarios with different client scales: total 10 clients with 100\% participation and total 100 clients with 10\% participation. 
We set the training parameters as 5 local epochs, the same number of communication rounds (20 rounds for the former and 100 rounds for the latter), and learning rates ($0.01$ for MNIST and FMNIST, $0.1$ for CIFAR-10). 
For the SV related hyper-parameters, we set the Monte Carlo sampling number as $R=3|S_{i,k}^t|$, where the number of model parameters downloaded for each round is $k=5$ in the beginning.
Note that $k$ is dynamically adjusted according to the dynamic top-$k$ download mechanism in \textit{SV for data relevance} of Sec. \ref{Sec:problem_solution_sv}.

\subsection{Performance Analysis}
\label{Performance Analysis}

In this section, we will demonstrate the performance of our pFedSV compared to all the state-of-the-art benchmarks and analyze the experiment results in details.

\textbf{Results on the different Non-IID data setting.}
Table 1 demonstrate the MTA of all methods with the pathological Non-IID data setting. 
Since each client has only two types of labels, which significantly simplifies the complexity of the classification task for each client, the high performance of separate on all datasets reflects the simplicity. 
However, the pathological Non-IID data setting is a great challenge for the single global model methods, we can observe that FedAvg and FedProx suffer from significant performance degradation on all datasets, since its global aggregation will contain models of data-irrelevant clients and thus lead to severe instability in the gradient optimization process \cite{zhang2020fedpd}.

For the other PFL methods, FedAvg+FT, pFedMe, FedFomo, FedAMP, pFedHN and our pFedSV all realize a promising performance on all datasets. FedAvg+FT takes several local fine-tuning steps to tune the poor global model back to adapt the local Non-IID data distribution. 
The pFedMe proposes novel regularized loss functions based on Moreau envelopes to decouple the personalized optimization from the global model learning. 
The pFedHN is more specific in that it directly generates personalized parameters for each client's model through another hypernetwork.
The good performance of FedFomo and FedAMP are achieved by adaptively encourage more pairwise collaboration between clients with similar models to form their own personalized model.
Our pFedSV can outperform all other baselines since it consider the multiwise influences among clients to help them identify their data-relevant coalition and generate personalized aggregation weights with multiwise collaboration.



Table 2 illustrates the MTA of all methods on the Dirichlet Non-IID data setting ($\alpha=0.1$). As we know from the visualization in Fig.~3,
this setting (Dirichlet $\alpha=0.1$) is much more challenge than pathological, which is reflected in the significant performance reduction of all methods.
Nevertheless, our pFedSV is still guaranteed to outperform all other baselines.


\textbf{Relevance score \& Multiwise collaboration weights.}
The superior performance of pFedSV comes from the various desirable properties of SV, 
Fig.~3 visualize the relevance vector $\phi^i$ of each client after convergence on different algorithms, where FedFomo in Fig.~3(b) use the model similarity based weights to update the relevance vector while pFedSV in Fig.~3(a) use the computed SV.
To illustrate the effectiveness of pFedSV, we also attach the visualized ground-truth of client relevance in Fig.~3(c) according to the client label distribution in Fig.~3(d) obtained from the omniscience perspective.
Obviously, it is evident from ground-truth that symmetry is an important feature of client relevance matrix, pFedSV perfectly identifies the real data-relevant clients and assigns aggregation weights with multiwise collaboration in the coalition, which significantly outperforms other pairwise collaboration methods based on model similarity.

\section{Discussion}
\label{discussion}

We note the communication overhead and privacy concern that may arise from the model downloading required for each client to perform local SV evaluation. In this section, we will discuss the effective solutions for these potential issues.

\textbf{Communication Overhead.}
To reduce the communication overhead of downloading other client models for local SV evaluation, except our dynamic download mechanism in \textit{SV for Data Relevance} of Sec. \ref{Sec:problem_solution_sv}, we further exploit the advantage of common representation between clients.
Specifically, we note that for some learning tasks (i.e., image classification and next-word-prediction), some low-dimensional common representation with similar functions can be shared, such as feature extraction. The parts that each client really need to personalize are their unique local heads, such as classifier \cite{collins2021exploiting}.
Therefore, we only need to download other clients' local heads to generate respective models, which can significantly reduce the communication overhead. Please refer to the Appendix.2 for more experiment details. 

\textbf{Model Privacy.} 
Although we can achieve anonymity by removing any information related to the client's identity from the downloaded model parameters during the entire process of pFedSV, it still may pose privacy issues in real-world scenarios. 
Therefore, we consider to adopt the $(\epsilon,\delta)$-differential privacy (DP) to address the privacy concerns \cite{abadi2016deep}. 
We can add Gaussian noise into the model parameters after client's local training process, which can guarantee the model with DP.
The additional Gaussian noise can make the model private but at the cost of performance degradation. Please refer to the Appendix.3 for more experiment details. 
 \vspace{-5pt}


\section{Conclusion}
\label{conclusion}

In this paper, we focus on the fact that model aggregation is a collaboration process within the coalition and first introduce SV from coalition game theory, to analyze the multiwise influences among clients and quantify their marginal contribution to the personalization of others. 
We propose a novel algorithm pFedSV to simultaneously address two critical challenges in PFL: identifying the optimal collaborator coalition and generating personalized aggregation weights based on multiwise influences analysis.
We conducted extensive experiments to demonstrate the effectiveness of pfedSV and the results empirically illustrate the superiority of multiwise collaboration through the significant improvement on the personalized accuracy.

\bibliographystyle{named}
\bibliography{ijcai21}

\end{document}